\documentclass[runningheads]{llncs}
\usepackage{tikz}
\usepackage{indentfirst}
\usepackage{float}
\usepackage{multicol}
\usepackage{svg}
\usepackage{cite}
\usepackage{amsmath,amssymb,amsfonts}
\usepackage{algorithmic}
\usepackage{graphicx}
\usepackage{textcomp}
\usepackage{algorithm}
\usepackage{graphicx}
\usepackage{subcaption}
\usepackage{pgfplots} 
\pgfplotsset{compat=newest} 
\usepackage{array}
\usepackage{tabularx}
\newcolumntype{Y}{>{\centering\arraybackslash}X}

\definecolor{fluo}{RGB}{204,255,0} 
 
\usepackage{threeparttable} 

\begin{document}
\title{Efficient Reinforcement Learning via Decoupling
Exploration and Utilization}
%
%
\author{
    Jingpu Yang\inst{1} \and
    Helin Wang\inst{1} \and
    Qirui Zhao\inst{1} \and
    Zhecheng Shi\inst{1} \and
    Zirui Song\inst{2} \and
    Miao Fang\inst{1,3}\thanks{Corresponding author.}
}
\institute{
    Northeastern University, Shenyang, China \and
    University of Technology Sydney, Sydney, Australia \and
    Olimei Company, Guangzhou, China
}

\maketitle              
\begin{abstract}
Reinforcement Learning (RL), recognized as an efficient learning approach, has achieved remarkable success across multiple fields and applications, including gaming, robotics, and autonomous vehicles. Classical single-agent reinforcement learning grapples with the imbalance of exploration and exploitation as well as limited generalization abilities. This methodology frequently leads to algorithms settling for suboptimal solutions that are tailored only to specific datasets.
In this work, our aim is to train agent with efficient learning by decoupling exploration and utilization, so that agent can escaping the conundrum of suboptimal Solutions.
In reinforcement learning, the previously imposed pessimistic punitive measures have deprived the model of its exploratory potential, resulting in diminished exploration capabilities. To address this, we have introduced an additional optimistic Actor to enhance the model's exploration ability, while employing a more constrained pessimistic Actor for performance evaluation. The above idea is implemented in the proposed OPARL (Optimistic and Pessimistic Actor Reinforcement Learning) algorithm.
This unique amalgamation within the reinforcement learning paradigm fosters a more balanced and efficient approach. It facilitates the optimization of policies that concentrate on high-reward actions via pessimistic exploitation strategies while concurrently ensuring extensive state coverage through optimistic exploration. Empirical and theoretical investigations demonstrate that OPARL enhances agent capabilities in both utilization and exploration. In the most tasks of DMControl benchmark and Mujoco environment, OPARL performed better than state-of-the-art methods.

\end{abstract}





\section{Introduction}

Reinforcement Learning(RL)\cite{servey}, either in single-agent\cite{success1,success3,game} or multi-agent\cite{wang2022robust,yuan2022multi,zhang2022discovering,yuan2023survey,yuan2023robust} contexts, has garnered considerable attention due to its potential for a variety of real-world applications. The latest research indicates that, Deep RL has achieved considerable success in various fields such as robotics\cite{robort}, recommendation systems\cite{recommendation}, and strategy games\cite{strategy/games}. In single-agent training, DQN\cite{DQN} surpassed human-level performance in Atari video games, while Alpha\cite{AlphaGo,AlphaGoZero,AlphaGoLee} defeated numerous professional players in the game of Go, achieving unprecedented heights.However, there are some problems in the generalization of single-agent.
Due to the divergence in data distribution between the training and testing environments, conventional RL algorithms are significantly impacted by extrapolation error, resulting in the overestimation of Q-values for Out-Of-Distribution (OOD) state-action pairs\cite{OOD1}. Overestimation can make certain state-action pairs appear more attractive than their actual value warrants. This leads to the agent erroneously prioritizing these overestimated actions, while neglecting other potentially superior but underestimated actions. Furthermore, the imbalance between exploration and exploitation strategies can lead to diminished model generalization capabilities.

\begin{figure}[htbp]
    \centering
    \vspace{-8mm} 
    \begin{subfigure} [b]{0.45\textwidth}
        \centering    
        \includegraphics[width=0.55\linewidth]{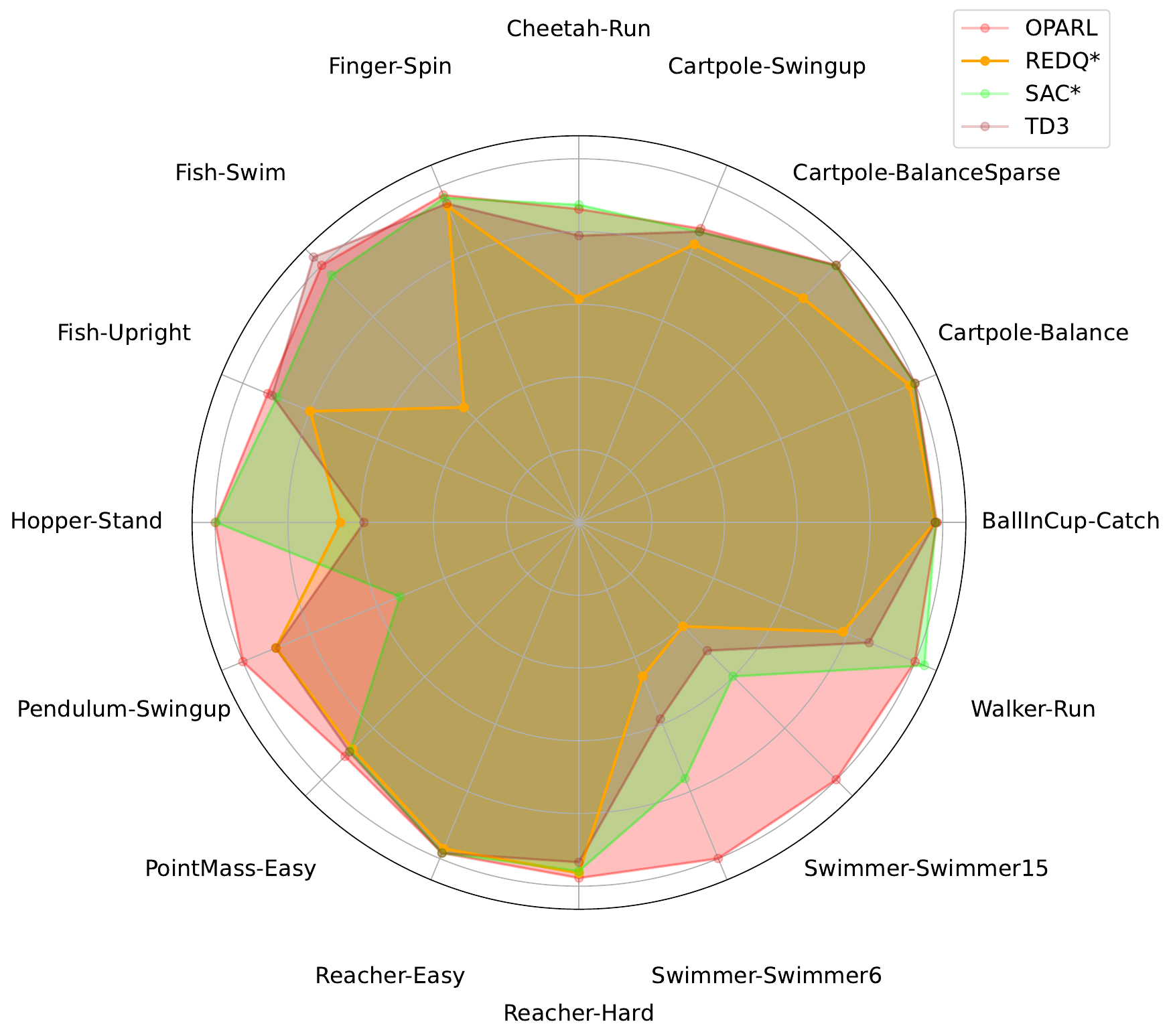}
        \caption{ Our model was tested for performance in the DMcontrol environment in comparison with other models.}
        \label{fig:alg}
    \end{subfigure}
    \hfill
    \begin{subfigure}[b]{0.45\textwidth}
        \includegraphics[width=0.55\linewidth]{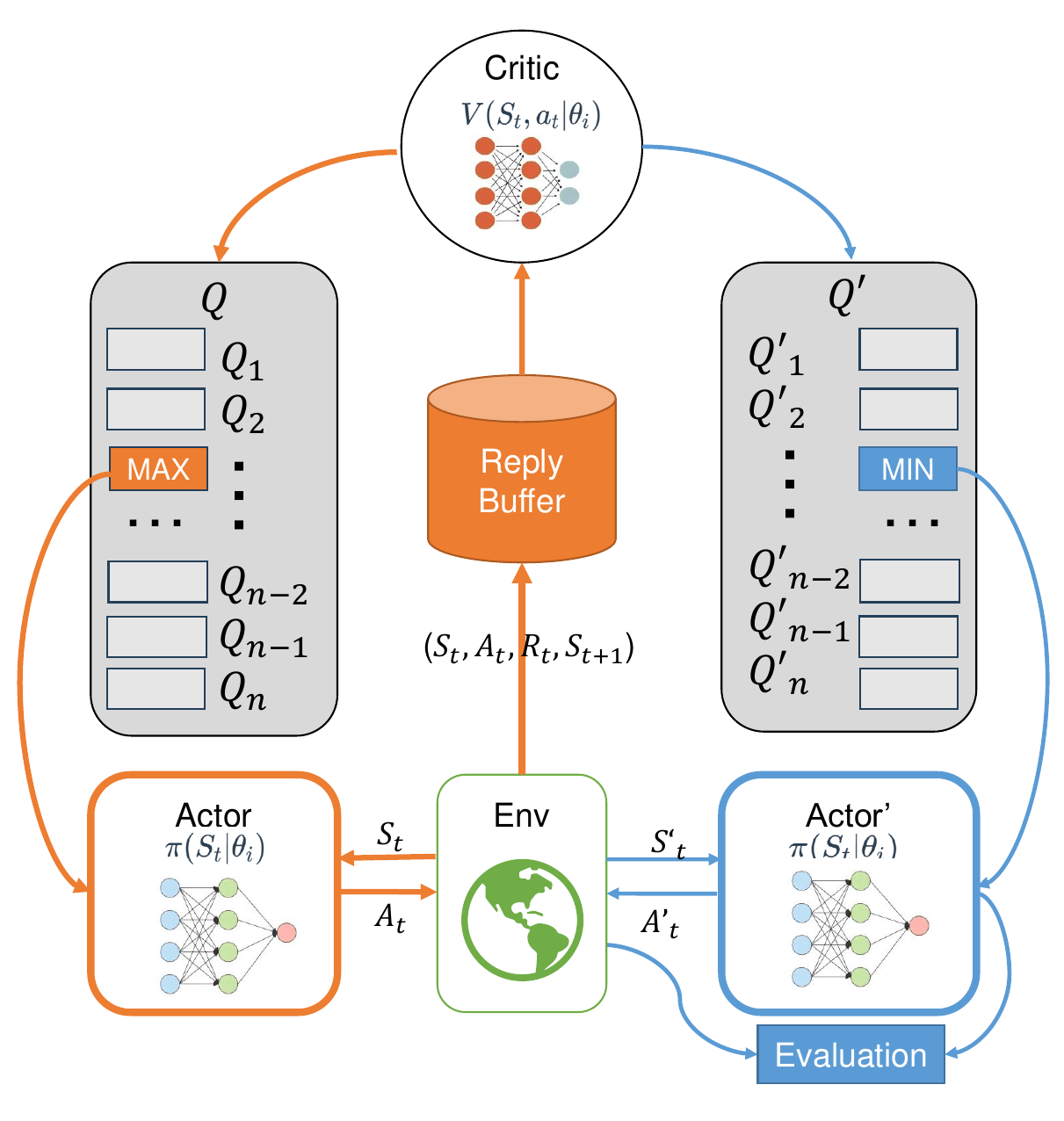}
        \caption{      Visualization of the OPARL framework.                            Blue represents pessimistic thoughts and orange represents optimistic thoughts.}
        \label{fig:fig2}  
        \label{fig:enter-label} 
    \end{subfigure}
\end{figure}

\vspace{-5mm}
To address these issues, previous approaches introduced conservative ideas to alleviate these problems, such as behavior cloning\cite{TD3BC} or forced pessimism. Behavior Cloning learns policies directly from expert behavior data, rather than through interaction with the environment. This results in the model being reliant on a large volume of high-quality expert data and potentially incapable of making correct decisions when encountering novel states not present in the training set, thus leading to limited generalization capabilities of the model. Trust Region Deep Deterministic Policy Gradient(TD3) trains two independent Q-networks to independently estimate the Q-values for the same state-action pairs\cite{TD3}, employing a pessimistic approach by selecting the lower of the two estimates for the same state-action pair. This method of reducing overestimation enhances the algorithm's stability and performance. However, enforced pessimism may lead to the underestimation of certain advantageous state-action pairs, thereby constraining the policy's potential for exploration.
Consequently, prior research has focused on the pessimistic constraint of optimistic exploration under uncertainty, which implies that overestimation of expected rewards could initiate exploration of states and actions that would otherwise remain unexplored. Thus, we engage in exploring the general form of the principle of optimistic uncertainty—overestimation of expected rewards—to stimulate the investigation of states and actions typically not explored while simultaneously ensuring the model's performance and stability.

In the realm of Reinforcement Learning for continuous control problems, two roughly antithetical methodologies have traditionally prevailed. On one hand, several researchers endeavor to mitigate overestimation bias, for instance, by adopting the minimum of the estimated values or by employing multiple estimates to formulate an approximate lower bound, thereby manifesting a form of pessimism towards the current value function estimation. Conversely, there exists a school of thought advocating that an inherent optimism in approximation estimations effectively fosters exploration within the environment and action space, thereby augmenting the potential for higher reward acquisition. Both approaches have underpinned the development of state-of-the-art algorithms, as evidenced by the literature \cite{leguantansuo, leguan, beiguan}, suggesting a non-mutually exclusive relationship between these methodologies. Consequently, our study aims to amalgamate these seemingly disparate strategies. To this end, we introduce the conceptualization of the OPARL (Optimistic and Pessimistic Actor Reinforcement Learning) algorithm, which bifurcates the role of the actor in the Actor-Critic model into two distinct components: one dedicated to optimistic exploration and the other to pessimistic exploitation. During the exploration phase, our model employs a highly optimistic state overlay, thereby enhancing the probability of identifying regions of high reward. Nonetheless, such a technique may not consistently converge to the optimal strategy. In contrast, during the policy recuperation phase, we implement strategic pessimism to optimize the exploitation of previously recognized high-reward behaviors. The overall concept is summarized in Figure \ref{fig:fig2}.

Our series of experiments have demonstrated that OPARL can balance both pessimistic and optimistic algorithms, offering a wider range of exploration and more stable performance. Our research validates that conducting exploration and training in an environment inclined towards highly optimistic estimations, particularly characterized by an overestimation of expected rewards, is not only feasible but also effective. Simultaneously, by decoupling and rebalancing exploration and exploitation, superior results are achieved. The experiments indicate that OPARL effectively enhances the model's generalization ability. Our model demonstrates improved states and rewards in four Mujoco environments and 14 out of 18 DMControl environment, The effect is as shown in the Figure \ref{fig:alg}.

\section{RELATED WORK}

\subsection{Reinforcement Learning}

In the realm of reinforcement learning\cite{qhxxtiaozhan}, a machine learning approach characterized by interaction with the environment, significant advantages have been demonstrated in diverse domains such as robotics control\cite{robort}\cite{robort1}, game theory\cite{game}\cite{strategy/games}, and autonomous vehicles\cite{zidongjiashi}. Within these applications, agents engage in a trial-and-error methodology to learn optimal action strategies, thereby achieving peak performance in specific settings. Reinforcement learning can be categorized into online and offline modalities. In online reinforcement learning, agents interact with the environment in real-time to learn strategies. However, this method's primary drawback lies in its reliance on immediate environmental feedback, which can slow the learning process and pose safety risks in real-world applications. Conversely, offline reinforcement learning utilizes pre-collected data for agent training, reducing dependence on real-time environmental interactions but potentially limiting the strategy's generalization capabilities in the face of novel data.

In single-agent reinforcement learning, key challenges encountered during training include balancing exploration and exploitation, reward delays, and environmental uncertainty. Previous researchers have addressed these issues through various methods, such as employing advanced exploration strategies\cite{xianjintansuocelv}, designing more intricate reward functions\cite{fuzajianglihanshu}, and implementing Model Predictive Control (MPC)\cite{MPC} to enhance the learning efficiency and adaptability of the agents. However, advanced exploration strategies do not fully ensure a balance between exploration and exploitation. Consequently, we propose the OPARL algorithm, which decouples exploration and exploitation strategies in single-agent scenarios. By training with distinct parameters for each strategy, OPARL reduces interference between exploration and exploitation, leading to improved learning outcomes.

\subsection{Sample Efficient Reinforcement Learning}
In the domain of RL, sample efficiency emerges as a critical metric for evaluating algorithmic performance\cite{sample}, especially in scenarios necessitating physical interactions, such as robotics\cite{robort}\cite{robort1} and autonomous driving\cite{zidongjiashi}. The primary goal of sample-efficient reinforcement learning is to expedite the learning process with minimal environmental interactions, thereby facilitating rapid and accurate decision-making. The quintessential challenge in sample-efficient RL lies in extracting the maximal amount of useful information from each environmental interaction, while concurrently maintaining the stability and generalizability of the algorithm.

To address this challenge, researchers have pioneered various methodologies. A prominent approach is Model-Based Reinforcement Learning (MBRL), which constructs an internal model of the environment to predict future states and rewards, thereby reducing the dependency on actual environmental samples. This approach has been effectively implemented in the research conducted by Kurutach et al\cite{2018}. Additionally, Transfer Learning and Meta-Learning play pivotal roles in enhancing sample efficiency. These techniques expedite the learning process in new tasks by leveraging knowledge acquired from previous tasks, thereby optimizing sample utilization. The potential of Meta-Learning in swiftly adapting to new environments was demonstrated in the studies by Finn et al\cite{2017}.

However, the implementation of these techniques often necessitates complex algorithmic design and fine-tuning. Furthermore, their efficacy is typically confined to specific tasks and may not generalize across diverse tasks or environments. In contrast, our proposed algorithm is comparatively straightforward, achieving equivalent or superior performance with fewer interactions. It rapidly learns from limited experiments, circumventing the high costs associated with data collection and potential safety risks.
\vspace{-3mm} 
\subsection{Optimistic and Pessimistic in Reinforcement Learning}

In the realm of exploratory learning research, the strategies of optimistic and pessimistic exploration have long been contentious\cite{optandpes,Im2022,ibarz2021train}. Scholars argue that in uncertain environments, it is necessary to employ algorithms based on optimistic principles (reward maximization) for exploring state-action pairs with high cognitive uncertainty\cite{leguanwucha,ibarz2021train,zhang2022review,stadler2023approximating,bruckner2020belief}, and such overestimation bias may not always be detrimental. In certain cases, errors due to underestimation bias can be harmful, as overestimation bias aids in encouraging exploration\cite{leguantansuo}\cite{leguantansuo1} of actions that are overvalued, while underestimation bias may impede exploration. When these highly random regions correspond to areas of high value, promoting exploration is beneficial, yet underestimation bias could prevent agents from learning the high values of such regions. Consequently, our algorithm retains a highly optimistic approach in exploration strategies to probe more high-value areas.

However, if highly random regions also possess lower values, overestimation bias might lead to agents excessively exploring low-value areas. Thus, another group of scholars emphasizes the necessity to standardize highly uncertain state-action pairs. Based on the DDPG\cite{DDPG} algorithm, the TD3\cite{TD3} method selects the smaller of two $Q$ values\cite{shuangQ}\cite{A3C} as an approximate lower bound and explores the environment pessimistically. By reducing pessimistic exploration in value estimation, agents can achieve better long-term rewards. Therefore, our algorithm maintains this pessimistic approach in utilization strategies to minimize extrapolation errors, ensuring model stability and performance.

\section{PROBLEM SETUP}
We will formulate the problem as a Markov Decision Process (MDP) \( M \equiv (S, A, R, p, \gamma\)), where:

\begin{itemize}
    \item \( S \) represents the state space,
    \item \( A \) represents the action space,
    \item \( p \) represents the transition dynamics,
    \item \( R \) represents the reward function,
    \item \( \gamma \in [0, 1) \) represents the discount factor.
\end{itemize}

For a given state \( s \in S \), the policy \( \pi \) maps the state to an action (deterministic policy), and the agent selects the action \( a \in A \) based on policy \( \pi \), receiving a reward \( r \) and a new state \( s' \) from the environment. Our goal is to learn behavior that maximizes rewards, where the benefit is defined as the total discounted reward \( R_t = \sum_{i=t}^{T} \gamma^{i-t} r(s_i, a_i) \), and the state-action value function \( Q_\pi(s, a) = \mathbb{E}_\pi \left[ \sum_{t=0}^{\infty} \gamma^t r(s_t, a_t) \mid s_0=s, a_0=a \right] \).

To learn behavior that maximizes rewards, we first need to obtain an experience pool buffer containing high-reward regions. The buffer is \( \{(s, a, s', r, d\_b \)\}, which is collected using the tactically optimistic behavior strategy \( \pi_{\text{opt}} \). The optimistic action \( a_{\text{opt}} \) is obtained through \( \pi_{\text{opt}} \), and a new state \( s' \) is obtained by interacting with the environment. The reward $r$ and whether the action is completed \(d\_b\) are added to the experience pool to obtain more high-reward regions. When using data from the experience pool, we extract states \( s \), actions \( a \), new states \( s' \), rewards \( r \), and whether the action is completed \(d\_b\) from the experience pool and use tactical pessimism to obtain more reasonable actions, thereby maximizing the reward.
\section{Optimistic and Pessimistic Actors in RL}
In section 4.1, we introduce the general OPARL framework, which can be effectively combined with any reinforcement learning algorithm for enhancement. In section 4.2, we delve into a mathematical exploration of merging optimistic exploration with pessimistic exploitation, illustrating how this strategy enables an agent to attain high rewards in the face of challenging exploration tasks. In section 4.3 explores the implementation of combining optimistic exploration and pessimistic exploitation through parameter resetting, demonstrating the synergy of these approaches.
\begin{algorithm}[H]
\caption{OPTIMISTIC AND PESSIMISTIC ACTOR IN RL (OPARL)}
\begin{algorithmic}[1] 
\REQUIRE Initialize critic networks $Q_{\theta 1}$, $Q_{\theta 2}$, \ldots, $Q_{\theta n}$, and actor network $\pi_{\phi opt}$, $\pi_{\phi pes}$; with random parameters $\theta_1$, $\theta_2$, \ldots, $\theta_n$, $\phi_{opt}$, $\phi_{pes}$;
\REQUIRE Initialize target networks $\theta_1' \leftarrow \theta_1$, $\theta_2' \leftarrow \theta_2$, \ldots, $\theta_N' \leftarrow \theta_N$, $\phi_{opt}' \leftarrow \phi_{opt}$, $\phi_{pes}' \longleftarrow \phi_{pes}$;
\REQUIRE Initialize replay buffer $\mathcal B$;\\
\renewcommand{\algorithmicrequire}{\textbf{Collect Data:}}
\REQUIRE 
\FOR{$t = 1$ \textbf{to} $T$}
            \STATE $a_{tu} = \pi_{\phi opt}(s_t) + \epsilon, \epsilon \sim N(0, s^2)$;
            \STATE Obtain new state $s_{t+1}$ and reward $r$;
            \STATE $Q_{max} = \max_{i \in N}(Q_{\theta i}(s_{t+1}, a_{tu}))$;
        \STATE $a_{opt} = a_{tu}$ where $Q_u = Q_{max}$;
    \STATE Sample mini-batch: $(s,a,r,s_{t+1})$ $\sim$ $\mathcal B$;\\
\renewcommand{\algorithmicrequire}{\textbf{Train Parameters:}}
\REQUIRE 
    \STATE $\tilde{a_{t}} = \pi_{\phi pes}(s_{t+1}) + \epsilon, \epsilon \sim N(0, s^2)$;
    \STATE Obtain new state $s_{t+1}$ and reward $r$;
    \STATE $Q_{min} = \min_{i \in N}(Q_{\theta i}(s_{t+1}, \tilde{a_{t}}))$;
    \STATE $a_{pes} = \tilde{a_{t}}$ where $Q_{\theta i} = Q_{min}$;
    \STATE Update critics:\\
    $y $ = $ r+\gamma \min_{i \in N} Q_{\theta i}(s', a_{pes})$;\\
    $\theta_i \longleftarrow \operatorname*{argmin}_{\theta_i}N^{-1}\sum(y-Q_{\theta_i}(s,a))^2$;
        \STATE Update target networks:\\
         $\theta'_i\leftarrow\tau\theta_i+(1-\tau)\theta'_i$ , where $ i \in N$;\\
         $\phi'_{pes}\leftarrow\tau\phi_{pes}+(1-\tau)\phi'_{pes}$;
    \IF{$t$ mod $w$}
        \STATE $\phi_{pes}\leftarrow \phi_{opt}$;
    \ENDIF
\ENDFOR

\end{algorithmic}
\label{algorithm:1}
\end{algorithm}

\subsection{The OPARL Framework}
Our algorithm is divided into two distinct phases: optimistic exploration and pessimistic utilization. During the exploration phase, state-action pairs with the maximum Q-value are determined through an optimistic strategy, and rewards $r$ and subsequent states $s_{t+1}$ obtained from interactions with the environment are stored in a buffer. Conversely, during the evaluation phase, we select state-action pairs with the minimum Q-value to compute losses and update parameters. Parameters from the optimistic strategy are periodically transferred to the pessimistic strategy function every $w$ steps, fostering a deeper integration of optimistic and pessimistic approaches. The pseudocode for OPARL is illustrated in Algorithm \ref{algorithm:1}.

\subsection{The Algorithm of OPARL}

We assume $Q_{\theta}'$ stands for optimistic exploration, while $Q_{\theta}''$ stands for pessimistic exploitation.
Formulas \ref{1}, \ref{2}, represent the exploration ranges that we consider to be pessimistic, optimistic.
\begin{equation}\label{1}
y1 = r + \gamma \min_{i = 1,2,\ldots,N} Q_{\theta_i}'(s', \widetilde{a}).
\end{equation}
\vspace{-2em} 

\begin{equation}\label{2}
y2 = r + \gamma \max_{i = 1,2,\ldots,N} Q_{\theta_i}''(s', \widetilde{a}).
\end{equation}
\vspace{-2em} 
\\

And Formula \ref{3} stands for random walk determined $Q^*_{\theta i}$ values from 1 to $N$, respectively. From this, we can discern that Formula \ref{3}, which involves random selection of state-action pairs among N $Q$-values, encapsulates exploration bounds that correspond to optimistic and pessimistic exploration at its upper and lower limits, respectively. This delineates the exploratory scope within which the random strategy operates, framed by the optimistic exploration's propensity for high-reward regions and the cautious approach characteristic of pessimistic exploration.

\begin{equation}\label{3}
y' = r + \gamma Q^*_{\theta i}(s', \widetilde{a}),\:where\:{i\in1,2,\ldots,N}.
\end{equation}
\vspace{-2em} 
\\

Formula \ref{4} suggests that the variance induced by pessimistic exploration is less than or equal to the variance from randomly selecting $Q$-values, implying a narrower exploratory range for pessimistic exploration. Conversely, Formula \ref{5} indicates that the variance caused by optimistic exploration is greater than or equal to the variance from random $Q$-value selection, suggesting a broader exploratory range for optimistic exploration. This academic interpretation aligns with the inherent nature of these exploration strategies within their respective scopes.

\begin{equation}\label{4}
\min_{\theta_{i}}N^{-1}\sum(y1-Q_{\theta_{i}}'(s,a))^2 \leq \min_{\theta_{i}}N^{-1}\sum(y'-Q_{\theta_{i}}'(s,a))^2.
\end{equation}
\vspace{-2em} 

\begin{equation}\label{5}
\max_{\theta_{i}}N^{-1}\sum(y2-Q_{\theta_{i}}'(s,a))^2 \leq \max_{\theta_{i}}N^{-1}\sum(y'-Q_{\theta_{i}}'(s,a))^2.
\end{equation}
\vspace{-2em} 
\\

Formula \ref{6} shows an estimate of future expectations. Therefore, as deduced from formulas \ref{5} and \ref{6}, optimistic exploration has the potential to explore more high-reward areas as it can engage in a greater number of activities while ensuring the expectations of those actions remain constant. Similarly, formulas \ref{4} and \ref{6} illustrate that despite unchanged expectations, the range of actions in pessimistic exploration undergoes a certain reduction. This implies that pessimistic exploration is inclined towards selecting reasonable actions from high-reward areas over those with the highest short-term returns. The OPARL algorithm effectively amalgamates optimism with pessimism, enabling broader exploration and achieving longer-term returns compared to other models.

\begin{align}\label{6}
Q_{\theta_{i}}(s_{t},a_{t}) &= r_{t} + \gamma \mathbb{E}[Q_{\theta}(s_{t+1},a_{t+1})].
\end{align}

\subsection{Parameter Reset}

In our investigation, we aim to delineate the parameters of a synergistic approach that integrates the characteristics of a pessimistic exploration strategy, denoted as $\phi_{pes}$, into the framework of an optimistic exploration approach, symbolized as $\phi_{opt}$. The principal objective of this integration is to empower the optimistic exploration strategy with the capacity to assimilate insights and knowledge gleaned from the pessimistic exploration paradigm, thereby augmenting its exploratory efficiency and effectiveness\cite{beguanleguanjiehe}. This methodological integration is operationalized by transferring the state dictionary from the pessimistic exploration strategy to the optimistic exploration framework. Such a transfer enables the initiation of the optimistic strategy with parameter configurations that bear resemblance to those employed by the pessimistic strategy. This strategic alignment is instrumental in mitigating the risk of excessive deviation from established exploration protocols and ensuring a grounded approach to exploratory behavior.

Concurrently, as the training process evolves, the optimistic exploration strategy is designed to dynamically refine its behavior policies through continuous parameter updates\cite{dongtaixingweijinglian}. This progressive adaptation fosters the development of more sophisticated and effective exploration techniques, thereby enhancing the strategy's capacity to proficiently execute its designated tasks. Additionally, this approach underscores the importance of maintaining congruence between the target policy network and the current policy network throughout the training phase. Such an alignment is pivotal for ensuring consistency and coherence in the training process\cite{celvewangluoyizhixing}. The versatility of this integrated approach renders it applicable for refining any exploration model that encapsulates elements of both optimism and pessimism in its operational paradigm.

\begin{figure*}[!htb]
\centering
\begin{tikzpicture}
    \begin{axis}[
        hide axis,
        xmin=10,
        xmax=50,
        ymin=0,
        ymax=0.4,
        legend style={draw=none,legend cell align=left},
        legend columns=-1
    ]
    \addlegendimage{red, thick}
    \addlegendentry{OPARL}
    \addlegendimage{blue, thick}
    \addlegendentry{TD3}
    \addlegendimage{green, thick}
    \addlegendentry{PPO}
    \addlegendimage{purple, thick}
    \addlegendentry{SAC}
    \addlegendimage{orange, thick}
    \addlegendentry{EBD}
    \addlegendimage{cyan, thick}
    \addlegendentry{RND}
    \end{axis}
\end{tikzpicture}
\vspace{0mm} 

\begin{subfigure}[b]{0.24\linewidth}
\centering
\includegraphics[width=\linewidth]{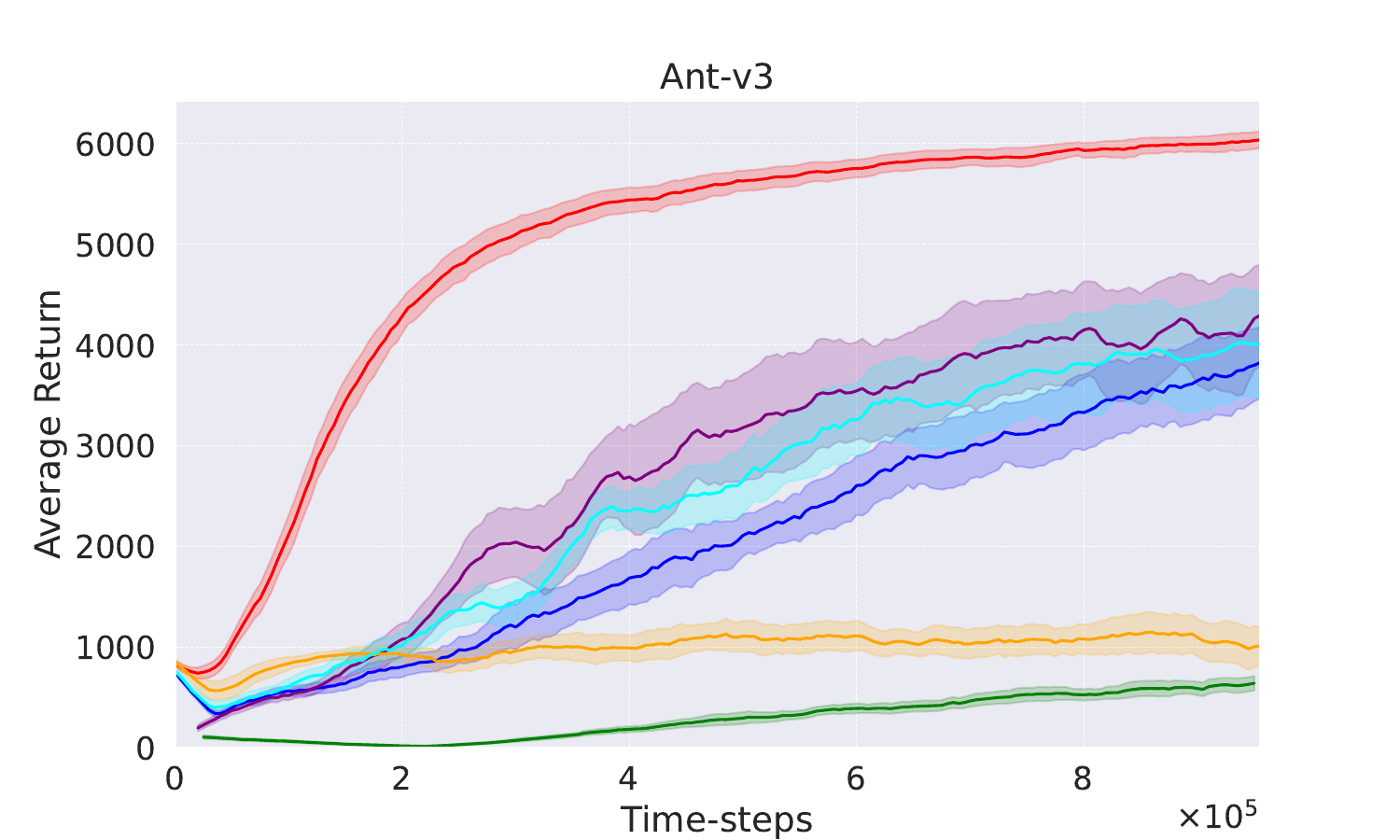}
\caption{Ant}

\end{subfigure}
\begin{subfigure}[b]{0.24\linewidth}
\centering
\includegraphics[width=\linewidth]{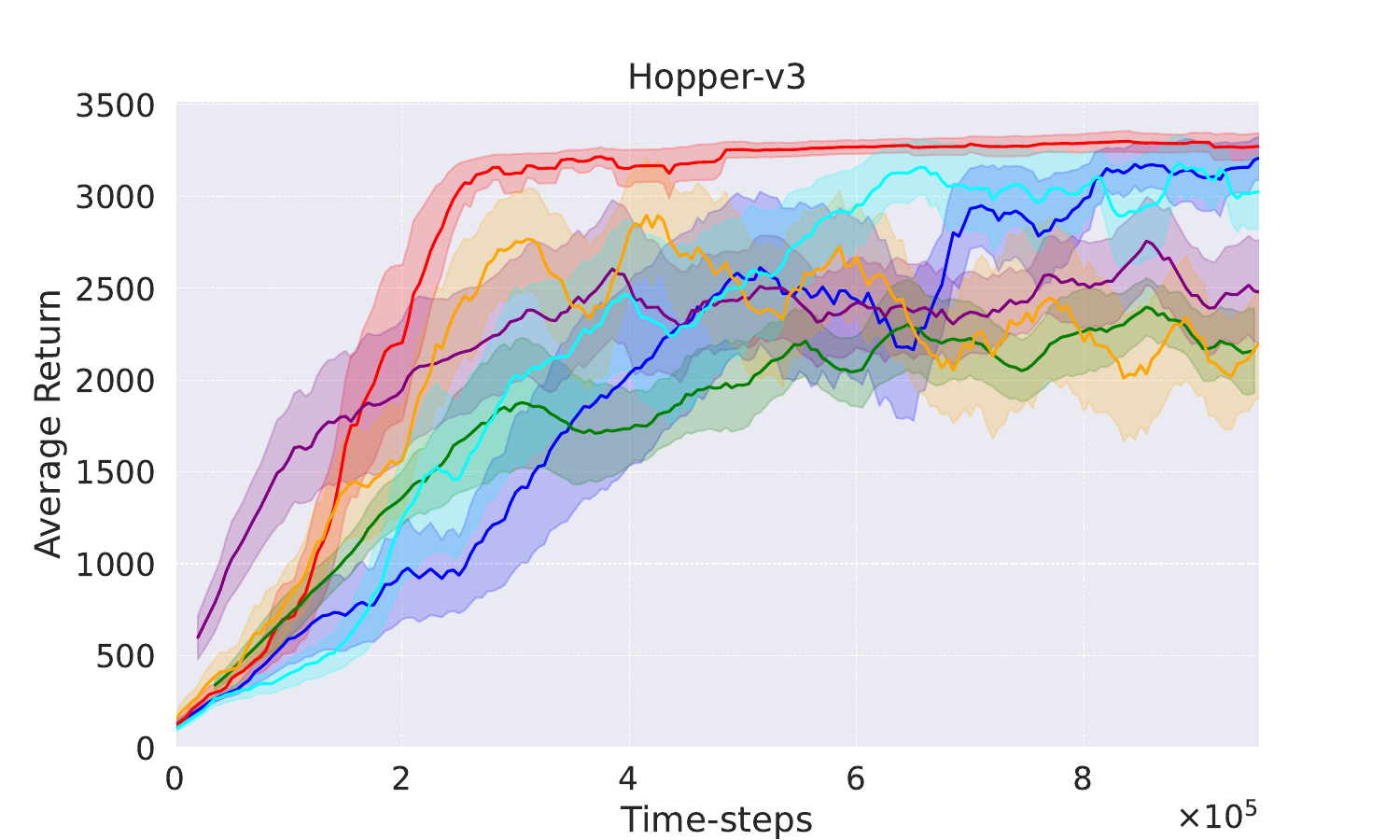}
\caption{Hopper}

\end{subfigure}
\begin{subfigure}[b]{0.24\linewidth}
\centering
\includegraphics[width=\linewidth]{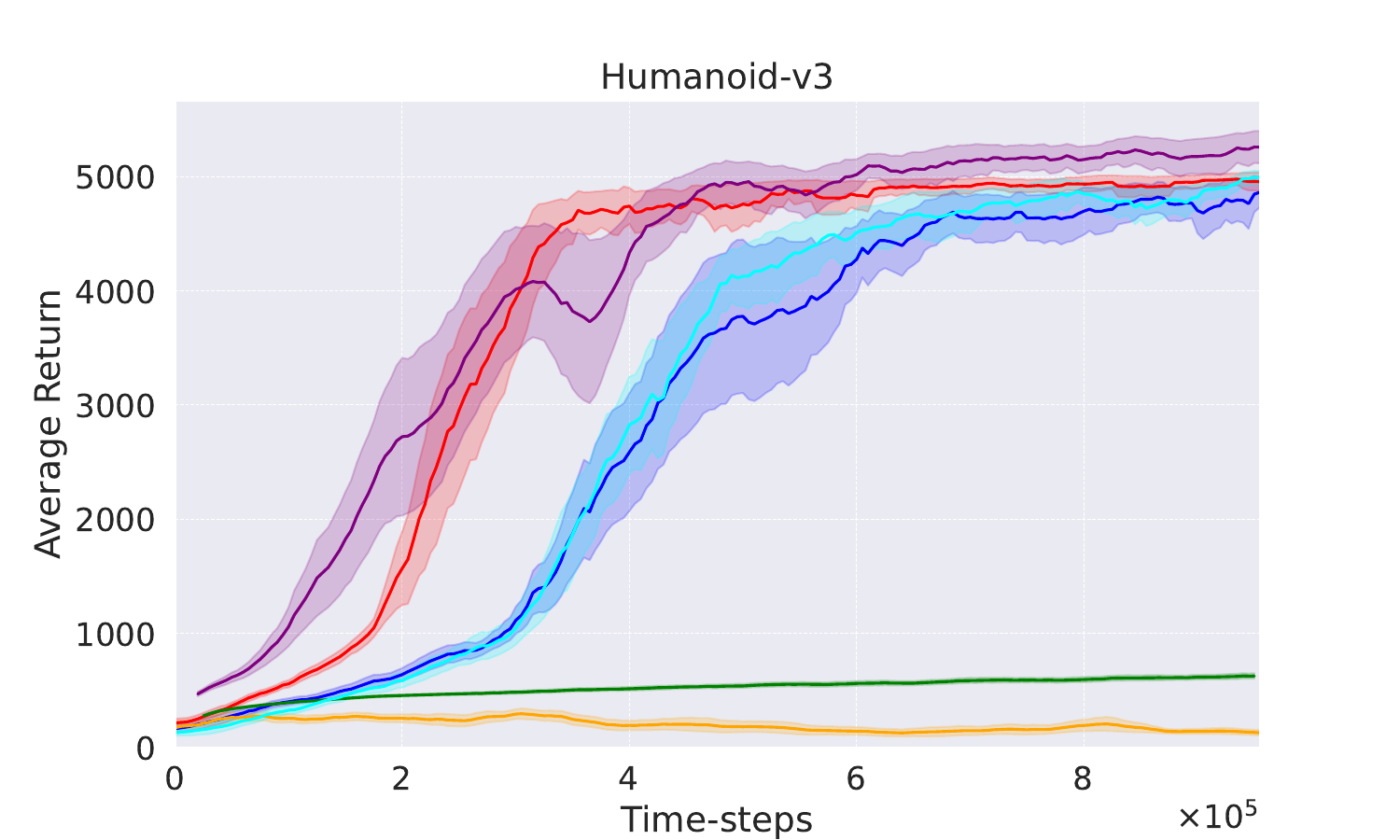}
\caption{Humanoid}

\end{subfigure}
\begin{subfigure}[b]{0.24\linewidth}
\centering
\includegraphics[width=\linewidth]{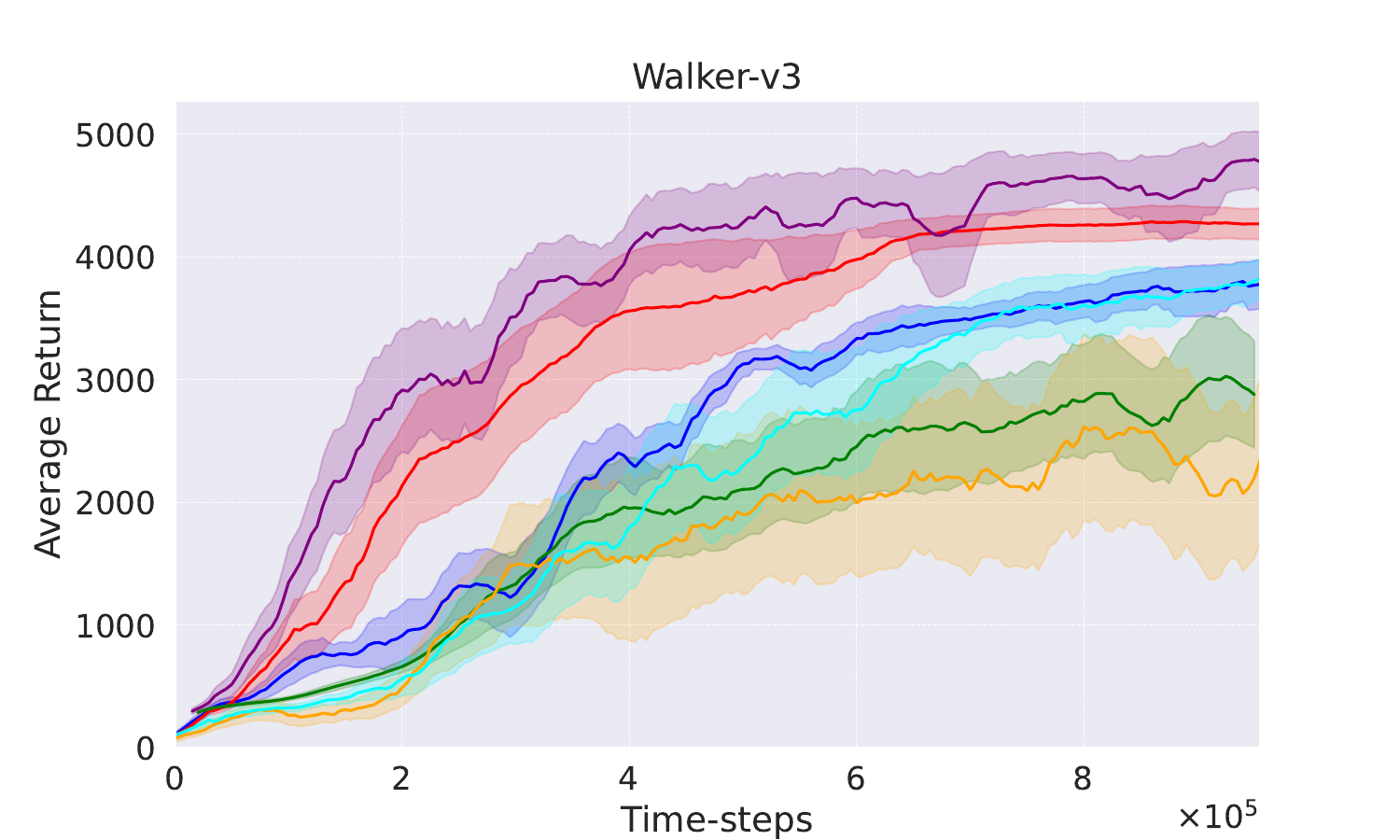}
\caption{Walker}

\end{subfigure}

\caption{
The performance curves presented in Mujoco environment, including
our OPARL and other benchmark algorithms within Mujoco framework. The shaded region represents half a standard deviation of the
average evaluation over 5 distinct seeds. Curves are smoothed with a window of size 10 for visual clarity.}
\label{fig:2}
\vspace{-6mm} 
\end{figure*}

\section{EXPERIMENTS}

Our experiments demonstrate the efficacy of decoupling exploration and exploitation strategies across a spectrum of tasks with varying demands for exploration. In Section 5.1, we elucidate the baseline and environmental prerequisites. In Section 5.2 provides a detailed comparative analysis of a range of online reinforcement learning algorithms within the Mujoco environment, juxtaposing their performance against our proposed model. In Section 5.3, the complexity is heightened by assessing our proposed OPARL algorithm within the challenging DMControl environment\cite{dmcshuju}, benchmarking it against baseline models, and providing an exhaustive analysis. In Section 5.4 delves into a series of ablation studies designed to substantiate the merits of decoupled exploration and exploitation. 
\vspace{-8mm}
\begin{table*}[h]
\caption{After undergoing training for one million time steps within the DMControl suite, the OPARL algorithm demonstrated superior performance in the majority of tasks. The highest scores are denoted in bold black typeface. The symbol ± represents the standard deviation across 10 trials.}
\begin{center}

\label{table:1}
\tiny 
\begin{tabularx}{1\textwidth}{YY|Y|Y|Y|Y}
\hline
\textbf{Domain} & \textbf{Task} & \textbf{OPARL} & \textbf{REDQ*} & \textbf{SAC*} & \textbf{TD3} \\ \hline
BallInCup & Catch & \textbf{983.8 $\pm$ 3.8} & \mbox{978.8 $\pm$ 3.7} & \mbox{980.3 $\pm$ 3.4} & \mbox{979.2 $\pm$ 1.0} \\
Cartpole & Balance & \textbf{\mbox{999.8 $\pm$ 0.0}} & \mbox{984.0 $\pm$ 6.0} & \mbox{997.7 $\pm$ 1.5} & \textbf{\mbox{999.8 $\pm$ 0.0}} \\
Cartpole & BalanceSparse & \textbf{\mbox{1000.0 $\pm$ 0.0}} & \mbox{872.1 $\pm$ 262.7} & \mbox{997.6 $\pm$ 5.7} & \textbf{\mbox{1000.0 $\pm$ 0.0}} \\
Cartpole & Swingup & \textbf{\mbox{874.6 $\pm$ 7.5}} & \mbox{828.1 $\pm$ 17.2} & \mbox{865.1 $\pm$ 1.6} & \mbox{865.9 $\pm$ 0.8} \\
Cheetah & Run & \mbox{861.6 $\pm$ 21.4} & \mbox{614.2 $\pm$ 58.2} & \textbf{\mbox{873.4 $\pm$ 21.5}} & \mbox{788.7 $\pm$ 50.9} \\
Finger & Spin & \textbf{\mbox{974.6 $\pm$ 15.5}} & \mbox{940.1 $\pm$ 33.5} & \mbox{966.3 $\pm$ 27.1} & \mbox{949.0 $\pm$ 16.2} \\
Fish & Swim & \mbox{356.0 $\pm$ 65.1} & \mbox{159.3 $\pm$ 100.1} & \mbox{342.4 $\pm$ 134.5} & \textbf{\mbox{367.4 $\pm$ 59.8}} \\
Fish & Upright & \textbf{\mbox{926.0 $\pm$ 18.6}} & \mbox{799.6 $\pm$ 113.8} & \mbox{898.6 $\pm$ 50.4} & \mbox{913.0 $\pm$ 14.6} \\
Hopper & Stand & \textbf{\mbox{599.5 $\pm$ 202.2}} & \mbox{393.5 $\pm$ 225.8} & \mbox{597.8 $\pm$ 308.8} & \mbox{354.4 $\pm$ 149.9} \\
Pendulum & Swingup & \textbf{\mbox{424.4 $\pm$ 155.0}} & \mbox{382.6 $\pm$ 297.0} & \mbox{226.2 $\pm$ 228.9} & \mbox{382.6 $\pm$ 209.3} \\
PointMass & Easy & \textbf{\mbox{909.4 $\pm$ 8.5}} & \mbox{880.9 $\pm$ 16.7} & \mbox{889.9 $\pm$ 33.1} & \mbox{892.5 $\pm$ 5.2} \\
Reacher & Easy & \mbox{983.6 $\pm$ 1.2} & \mbox{970.9 $\pm$ 24.4} & \mbox{983.5 $\pm$ 4.2} & \textbf{\mbox{983.8 $\pm$ 1.2}} \\
Reacher & Hard & \textbf{\mbox{977.5 $\pm$ 2.9}} & \mbox{964.1 $\pm$ 24.0} & \mbox{958.6 $\pm$ 40.9} & \mbox{934.1 $\pm$ 71.2} \\
Swimmer & Swimmer6 & \textbf{\mbox{471.4 $\pm$ 60.8}} & \mbox{215.8 $\pm$ 119.0} & \mbox{359.3 $\pm$ 130.9} & \mbox{275.6 $\pm$ 84.9} \\
Swimmer & Swimmer15 & \textbf{\mbox{442.1 $\pm$ 187.5}} & \mbox{178.6 $\pm$ 116.6} & \mbox{264.6 $\pm$ 136.9} & \mbox{220.4 $\pm$ 82.3} \\
Walker & Run & \mbox{752.1 $\pm$ 23.2} & \mbox{590.9 $\pm$ 51.6} & \textbf{\mbox{773.0 $\pm$ 32.9}} & \mbox{649.3 $\pm$ 82.9} \\
Walker & Stand & \textbf{\mbox{989.0 $\pm$ 2.0}} & \mbox{974.0 $\pm$ 12.6} & \mbox{986.8 $\pm$ 2.7} & \mbox{984.5 $\pm$ 1.3} \\
Walker & Walk & \textbf{\mbox{975.3 $\pm$ 2.9}} & \mbox{957.3 $\pm$ 10.6} & \mbox{973.5 $\pm$ 4.4} & \mbox{969.2 $\pm$ 2.1} \\
\hline
\textbf{Average} & \textbf{Scores} & \textbf{805.7} & 704.7 & 774.1 & 750.5 \\
\textbf{Improvements} & \textbf{Percentage} & / & 14.30\% & 4.08\% & 7.36\% \\
\hline
\end{tabularx}

 \begin{tablenotes}
        \footnotesize
        \item[*]The data marked with [*] is cited from the paper\cite{dmcshuju}
      \end{tablenotes}
\end{center}
\end{table*}

\vspace{-12mm} 
\subsection{Evaluation Setting}
\textbf{Baseline:}
In our research, we subjected the newly proposed OPARL algorithm to benchmark testing, assessing its efficacy relative to a cohort of acknowledged algorithms. Owing to their demonstrated reliability in various tasks, TD3\cite{TD3}, SAC\cite{SAC}, and Proximal Policy Optimization (PPO) \cite{PPO}were chosen as our principal benchmarks. In exploration comparisons, we selected state-of-the-art models such as RND\cite{rnd} and EBD\cite{ebd} for juxtaposition. Additionally, within the DMControl suite, we integrated comparisons with Randomized Ensembled Double Q-Learning (REDQ)\cite{REDQ}, which is distinguished by its ensemble Q-technology that augments exploratory capabilities and stability. To guarantee equitable and uniform comparisons, we utilized implementations from CleanRL with default hyperparameters as stipulated by the original authors.\\
\indent\textbf{Environments:} 
Our study utilized the state-based DMControl suite\cite{dmcshuju} under the Mujoco\cite{mujoco}\cite{mujoco1} framework provided by OpenAI's Gym\cite{GYM}, to assess performance across four continuous control tasks. The choice of OpenAI Gym\cite{GYM} was due to its array of environments conducive to benchmarking reinforcement learning algorithms, enabling efficient algorithm-environment interactions.\\
\indent\textbf{Setup:} 
We modulate the levels of optimism and pessimism in our model via the ensemble $Q$, initially set at five. The state-action pairs are determined by selecting the one with the greatest variance from these five $Q$-values. To balance optimistic and pessimistic explorations, we set their ratio at 1:1. Additionally, we revise the pessimistic exploration parameters $\phi_{pes}$ every 20000 time steps and incorporate them into the optimistic exploration parameters $\phi_{opt}$. The other parameter settings of this algorithm are in line with the original TD3 model\cite{TD3}. We believe that adhering to these baseline parameters facilitates a fair comparison with existing methodologies while allowing the nuances of our approach to be highlighted.

\subsection{Evaluation of OPARL in Mujoco Environment}

In the Mujoco environment, our results as depicted in the accompanying Figure \ref{fig:2}, reveal that the OPARL model exhibits exceptional performance, particularly in the Hopper and Ant scenarios. Notably, OPARL consistently surpasses its foundational algorithm, TD3, in the majority of test cases. A trend observed is the progressive stabilization of the model with increasing time steps, marked by OPARL exhibiting the least fluctuation amplitude among all models compared. This attribute underscores OPARL's capacity to enhance learning stability. In a comprehensive assessment, particularly in the Ant environment, our model significantly outperforms competing models. Quantitatively, OPARL achieves performance improvements of 44.04\%, 51.25\% and 773.02\% over the SAC, TD3, and PPO\cite{PPO} models, respectively. At the same time, our algorithm achieves performance enhancements of 50.93\% and 497.62\% over the RND\cite{rnd} and EBD\cite{ebd}. This significant improvement underscores the efficacy of integrating multiple Q-values and adopting a maximal value approach (optimistic exploration strategy) in the decision-making process, which markedly augments the algorithm's capacity for exploration. Consequently, this approach substantially broadens its adaptability and learning scope within complex environments.

\subsection{Evaluation of OPARL in DMControl Environment}

Table \ref{table:1} delineates the mean performance metrics of the algorithm after a training regimen spanning one million time steps. The empirical outcomes significantly indicate that the OPARL algorithm surpasses the comparative algorithms in the majority of the environments tested (14 out of 18). Specifically, when juxtaposed with the REDQ, SAC, and TD3 algorithms, OPARL demonstrated performance enhancements of 14.30\%, 4.08\%, and 7.36\%, respectively. Notably, in the "swimmer-swimmer6" and "swimmer-swimmer15" scenarios, OPARL's performance increments were particularly pronounced, amounting to 71.04\% and 100.59\%, respectively. These findings not only underscore the adaptability and superiority of OPARL across a diverse array of environments but also lay a solid groundwork for further exploration of its potential in complex task applications.

\subsection{Ablation Study}

\begin{figure*}[htbp]
\centering
\begin{subfigure}[b]{0.24\linewidth}
\centering
\includegraphics[width=\linewidth]{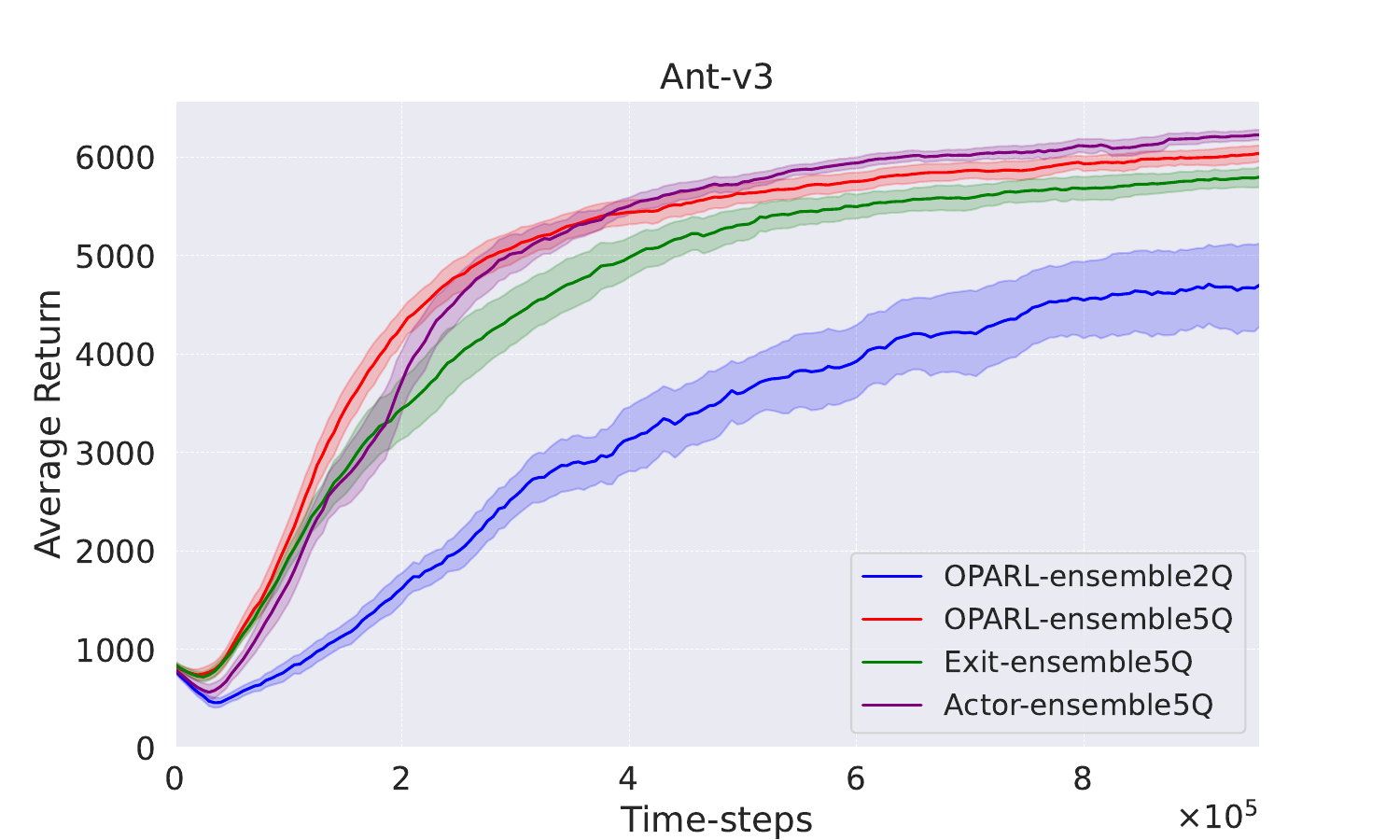}
\caption{Ant}

\end{subfigure}
\begin{subfigure}[b]{0.24\linewidth}
\centering
\includegraphics[width=\linewidth]{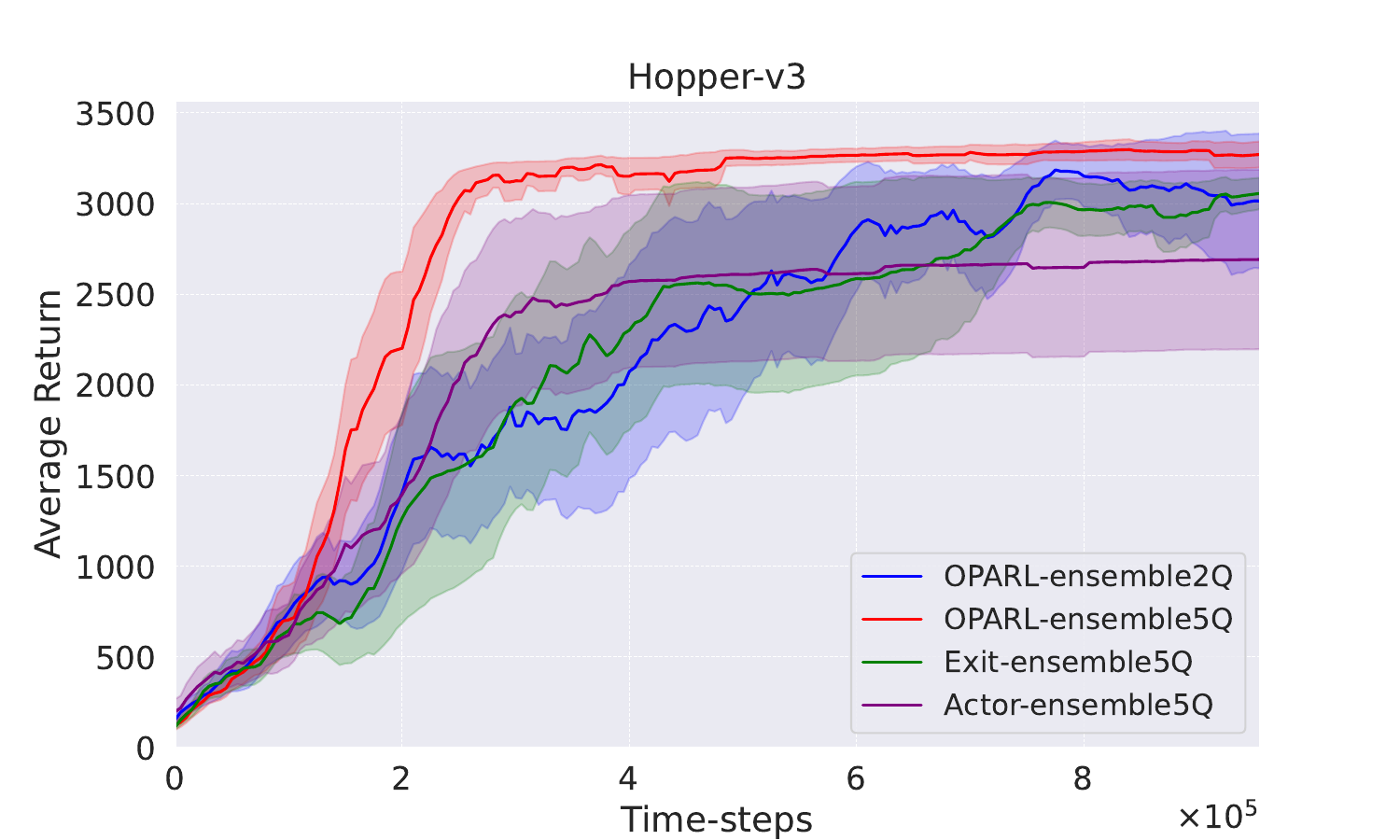}
\caption{Hopper}

\end{subfigure}
\hfill
\begin{subfigure}[b]{0.24\linewidth}
\centering
\includegraphics[width=\linewidth]{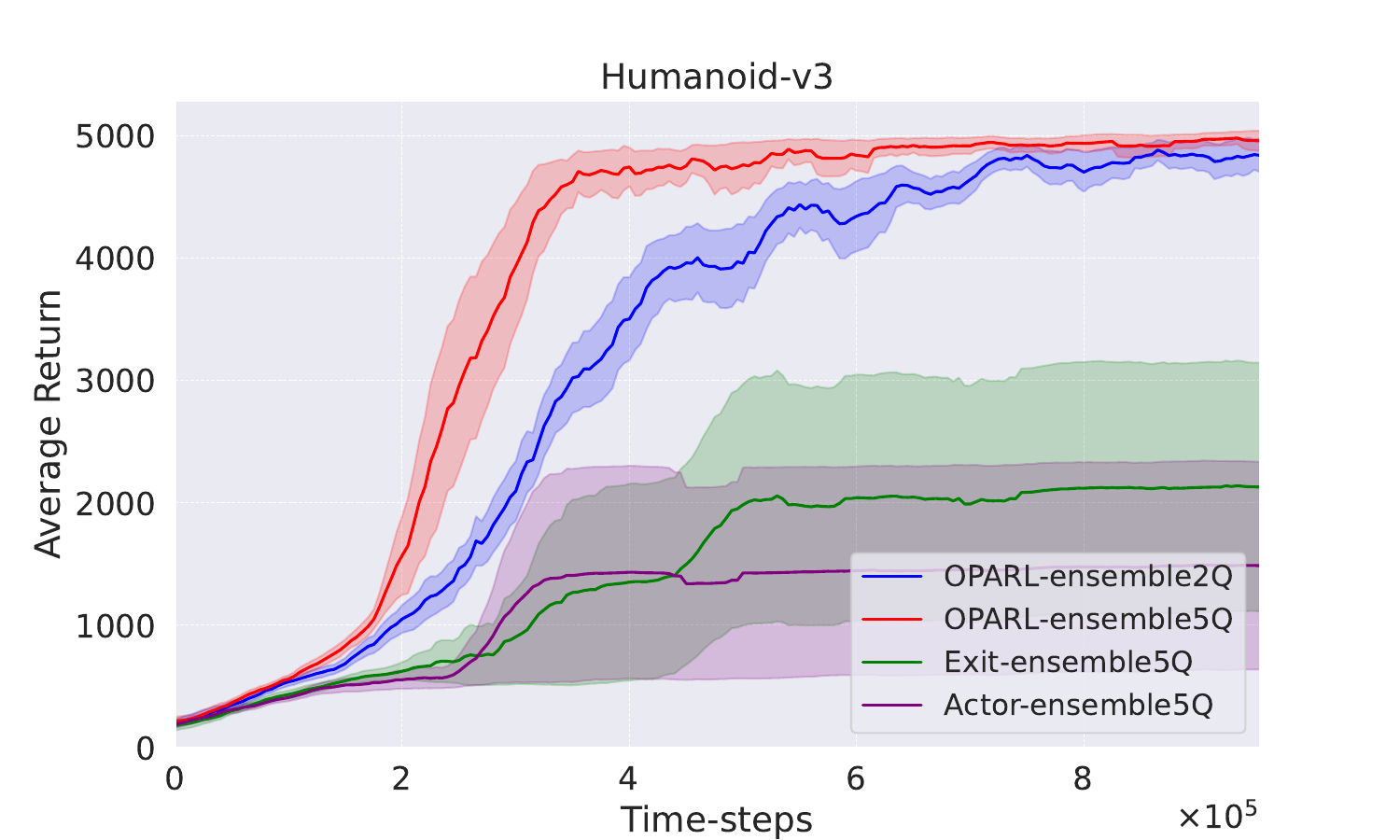}
\caption{Humanoid}

\end{subfigure}
\begin{subfigure}[b]{0.24\linewidth}
\centering
\includegraphics[width=\linewidth]{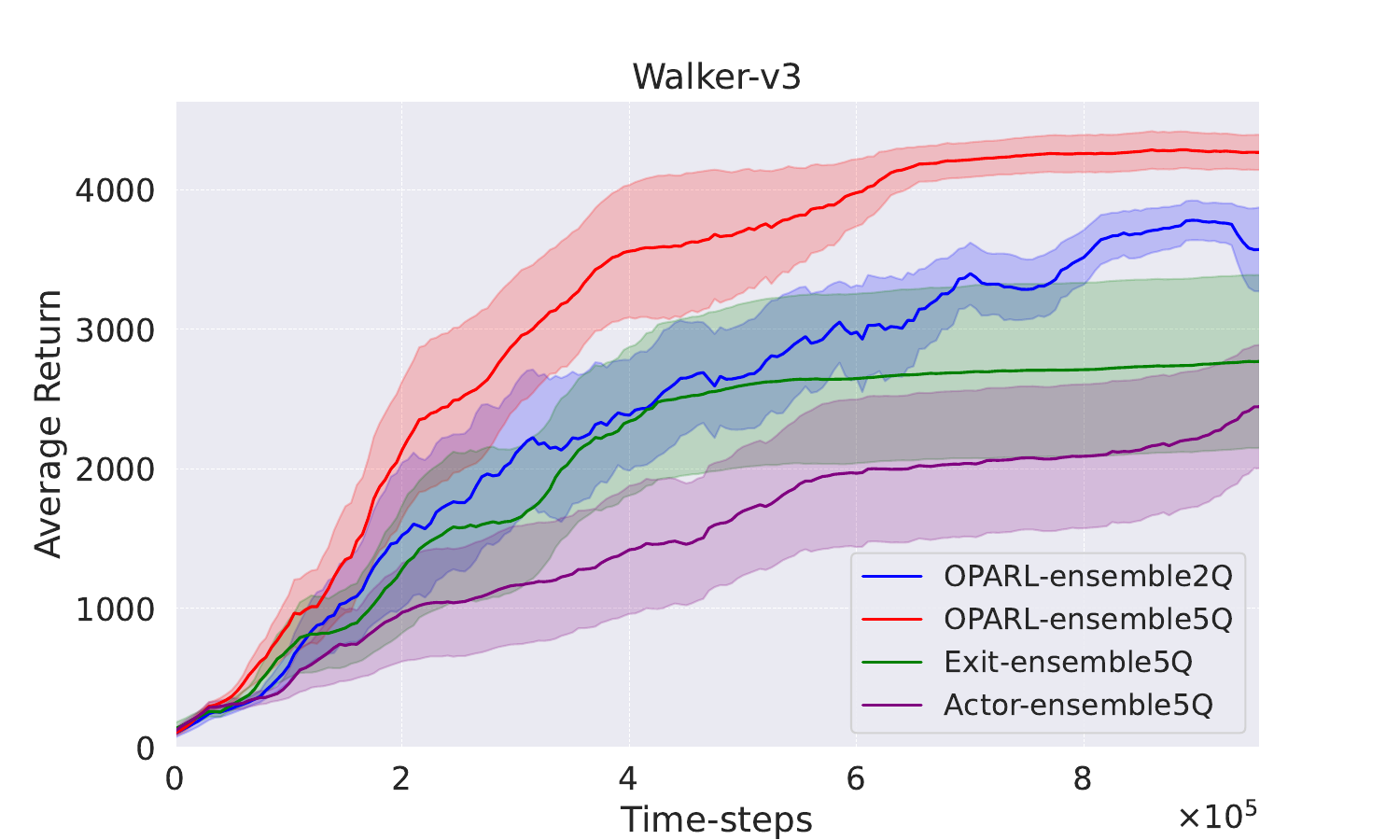}
\caption{Walker}

\end{subfigure}
\caption{The ablation performance curves presented in the Mujoco environment. The shaded region represents half a standard deviation of the average evaluation over 5 distinct seeds. Curves are smoothed with a window of size 10 for visual clarity.}
\label{fig:mujoco}
\end{figure*}

Figure \ref{fig:mujoco} presents a series of ablation studies aimed at evaluating the impact of novel components integrated into the architecture of our algorithm. The experimental design involved the independent application of an optimistic exploration strategy—selecting the maximal value from an ensemble of five $Q$ values—and a pessimistic exploration strategy—selecting the minimal value from the ensemble. Additionally, we examined the effect of not introducing extra ensemble $Q$ values, thus maintaining the original configuration with only two $Q$ values. The evaluation results are presented in the figure. While Yarats et al.'s study\cite{xiaorong1} suggests that adequate exploration and state coverage can enable standard reinforcement learning to achieve equivalent outcomes without a pessimistic strategy, both prior work on decoupled strategy learning\cite{xiaorong2} and our experimental findings consistently demonstrate that to enhance performance, optimistic exploration must be coupled with pessimistic constraints.

\vspace{-8mm} 
\section{CONCLUSION}

We propose Optimistic and Pessimistic Actor in reinforcement 
learning(OPARL), a straightforward and effective reinforcement learning algorithm that decouples the exploration and exploitation strategies of the model. This approach, characterized by optimistic exploration and pessimistic exploitation, achieves superior outcomes in most environments. Our research has demonstrated that high optimism in exploration is crucial in the learning process of complex environments. However, previous Actor-Critic algorithms, relying solely on a fixed degree of optimism, were unable to select the most rational actions from a long-term perspective, while traditional pessimistic trimming severely limited the model's exploratory capabilities. Our experiments show that by decoupling the original algorithm into distinct exploration and exploitation functions, OPARL significantly enhances performance in the Mujoco environment, especially achieving rapid learning in challenging settings such as Ant and Humanoid. Moreover, the model effectively adapts to the more complex DMControl environment, improving performance in 14 out of 18 settings, thereby reaching the state-of-the-art level.

\bibliographystyle{unsrt}
\bibliography{main}

\end{document}